\documentclass{article}

\usepackage{PRIMEarxiv}

\usepackage[utf8]{inputenc} 
\usepackage[T1]{fontenc}    
\usepackage{hyperref}       
\usepackage{url}            
\usepackage{booktabs}       
\usepackage{amsfonts}       
\usepackage{nicefrac}       
\usepackage{microtype}      
\usepackage{lipsum}
\usepackage{fancyhdr}       
\usepackage{graphicx}       
\usepackage{subfigure}
\usepackage{makecell}
\usepackage{natbib}
\graphicspath{{media/}}     

\usepackage[T1]{fontenc}
\usepackage[utf8]{inputenc}
\usepackage{authblk}

\title{NPU-BOLT: A Dataset for Bolt Object Detection in Natural Scene Images}
\author[1,2]{Yadian Zhao}
\author[1]{Zhenglin Yang}
\author[ 1,2]{Chao Xu \thanks{Corresponding author: chao\_xu@nwpu.edu.cn. Major field: Structural Health Monitoring, Structural Dynamics} }
\affil[1]{Schoolof Astronautics, Northwestern Polytechnical University, Xi'an 710072, China}
\affil[2]{Taicang Yangtze River Delta Research Institute, Northwestern Polytechnical University, Suzhou 215400, China}

\begin{document}
\maketitle

\begin{abstract}
Bolt joints are very common and important in engineering structures. Due to extreme service environment and load factors, bolts often get loose or even disengaged. To real-time or timely detect the loosed or disengaged bolts is an urgent need in practical engineering, which is critical to keep structural safety and service life. In recent years, many bolt loosening detection methods using deep learning and machine learning techniques have been proposed and are attracting more and more attention. However, most of these studies use bolt images captured in laboratory for deep leaning model training. The images are obtained in a well-controlled light, distance, and view angle conditions. Also, the bolted structures are well designed experimental structures with brand new bolts and the bolts are exposed without any shelter nearby. It is noted that in practical engineering, the above well controlled lab conditions are not easy realized and the real bolt images often have blur edges, oblique perspective, partial occlusion and indistinguishable colors etc., which make the trained models obtained in laboratory conditions loss their accuracy or fails. Therefore, the aim of this study is to develop a dataset named NPU-BOLT for bolt object detection in natural scene images and open it to researchers for public use and further development.  In the first version of the dataset, it contains 337 samples of bolt joints images mainly in the natural environment, with image data sizes ranging from 400*400 to 6000*4000, totaling approximately 1275 bolt targets. The bolt targets are annotated into four categories named blur bolt, bolt head, bolt nut and bolt side. The dataset is tested with advanced object detection models including yolov5, Faster-RCNN and CenterNet. The effectiveness of the dataset is validated. 
\end{abstract}


\section{Introduction}
Bolted joints are widely used in civil, ocean, aerospace and mechanical engineering fields for assembling parts and structures. Due to extreme service environment and loading factors, bolted loosening or even disengaging is often happened, which is a very concerned problem in engineering (\citet{1}).Timely detection of bolt loosening in the whole life duration of structure is urgently needed.
In recent years, the bolt loosening detection methods that using deep learning and machine learning techniques have been proposed and are attracting more and more attentions. However, most of these studies use bolt images captured in the laboratory for deep leaning model training, where the images are obtained in a well-controlled light, distance, and view angle conditions (\citet{2,3,4,5}). Also, the bolted structures are well designed experimental structures with brand new bolts and the bolts are exposed without any shelter nearby. It is noted that in practical engineering, the above well controlled lab conditions are not easy realized and the real bolt images often have blur edges, oblique perspective, partial occlusion and indistinguishable colors etc., which make the trained models obtained in laboratory conditions loss their accuracy or fails.
It is well known that datasets are very important for data-driven deep learning methods, especially in the field of target detection (\citet{6}). To largely promote the research on the bolt loosening detection method that using deep learning and machine learning technique and to make the method more practicality, it is necessary to develop a dataset with images from real natural scenes. Therefore, the aim of the work is to develop a dataset named NPU-BOLT for bolt object detection in natural scene images and open it to researchers for public use and further development. As far as the authors know, the dataset is the first public dataset dedicated to bolt object detection by using deep learning methods. Most of the images in NPU-BOLT dataset are captured in natural scenes by hand-hold cameras and drones, which try to cover as many possible situations in the natural environment as possible. Meanwhile, the dataset provided in this paper is uploaded publicly available on Kaggle platforms. The paper is organized as follows. In the second section, the composition of the dataset and its elements are briefly explained. In the third section, this paper explains the annotation method and classification basis of data set. In the fourth section, we select some advanced target detection model to test the datasets. Finally, in the fifth section, the work of this paper is summarized.

\section{NPU-BOLT dataset description}

The images in NPU-BOLT dataset are mainly from real images captured in natural environment. Specifically, these images mainly consist of two parts. One part is captured by the authors and the other part is downloaded from web. The first part includes 206 images and the second parts includes 114 images. Besides, there are a small number of bolt CAD simulation images. The dataset contains single bolt or double bolt images at close range, which tend to occupy a larger scale in the original image, as well as multi-bolt cluster images at a long distance.

\subsection{Images captured by the authors}
For the images captured by the authors in natural environment, the following practical conditions, different light intensity, shadow shielding, large area corrosion, bolt shedding, cracks, rain and snow weather are considered. The real images from natural scenes are captured by a Cannon 200D hand-hold camera, an Honor X10 mobile phone and a DJI AIR 2plus drone. Table 1 gives the specific parameters of the sampling device and the number of images captured by each device respectively. 

\begin{table}[ht]
 \caption{Specific parameters of the sampling device.}
  \centering
  \begin{tabular}{lll}
    \toprule
             & Image size	   & Image counts\\
    \midrule
    \centering
    DJI AIR 2plus (UAV)  & 5472*3648    &  \makecell[c]{18}    \\
    Canon 200D (camera)  & 6000*4000    &  \makecell[c]{21}    \\
    Honor X10 (phone)    & 2736*3648    &  \makecell[c]{167}  \\
    \bottomrule
  \end{tabular}
  \label{tab:table1}
\end{table}

In the dataset, the images of real bolt connections in sunny, cloudy, night and indoor (building) environments were collected. A variety of bolted structures are captured to cover as many as possible bolted joint types, which include bolted flange joints, pipe connections and different bolted structures from bridge steel frame, vehicle hub, construction machinery, guardrails, iron cable fixings and anchor bolts. The filename for these images captured by the authors are defined as AUT-X.
Figure 1 shows some typical captured images. It can be found the bolt images in the natural environment contain complex background information and bolts in the images often have corrosion and shelter. 

\begin{figure}[ht]
  \centering
  \includegraphics{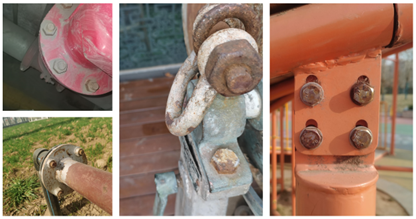}
  \caption{ Typical captured images in natural scene by the authors.}
  \label{fig:fig1}
\end{figure}

\subsection{Images downloaded from internet}

In this part, the bolt images are downloaded from internet to supplement the dataset. The reason is that if the images in the dataset are all from the same photographers, the photographers’ habits may reduce the diversity of the dataset. The images on the internet are generally captured by different photographers and covers many engineering objects. To include the images from internet can improve the application scope and diversity of the dataset. 
However, generally the pixel quality of these pictures is slightly lower than the pictures captured by the authors. And, most of the image sizes are between 500*500 and 1500*1500. The typical bolts images obtained from internet are shown in Fig 2. The filename for these images from web are defined as WEB-X.
\begin{figure}[ht]
  \centering
  \includegraphics{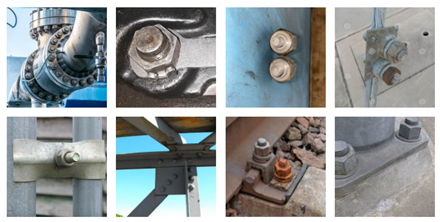}
  \caption{Typical Collected images from internet.}
  \label{fig:fig2}
\end{figure}

\subsection{3D CAD simulation images}

In this parts, the 3D CAD simulations images are added to the dataset. The reason is that 3D CAD images usually have high resolutions and clear bolt features without background information. Beside, the view angle can be easily adjusted in the CAD model. With adding a few number of 3D CAD simulation images in the dataset, it is expected to enhance the bolt object detection accuracy.  Specifically, total 17 CAD simulation images are included. The typical CAD images are shown in Figure 3. To distinguish these man-made CAD images from real bolt images obtained by the authors or internet, the filename for these images are defined as CAD-X.  The dataset user can choose whether to use this part for their training process. Figure 3 gives some typical CAD simulation images. 
\begin{figure}[ht]
  \centering
  \includegraphics{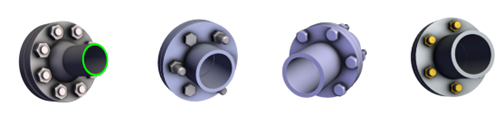}
  \caption{ Bolt images created by 3D CAD simulation.}
  \label{fig:fig3}
\end{figure}

\section{Annotation}
\subsection{Category}
NPU-BOLT dataset is specially annotated into the following four categories, which are bolt head, bolt nut, bolt side, and blur bolt. Due to the surround shelter and camera view, only some parts of a bolt are captured. The parts may be the side of the bolt, the nut of the bolt and the head of the bolt. In addition, due to the loss of camera focus, some bolts are blur in the captured images. Therefore, all captured images are divided into the above four categories. It is noted that to find a good camera angle which can avoid to capture the bolt side and blur samples is possible, but this is not easy for practical engineering applications. Figure 4 shows sample images for these four categories respectively. In figure 5, 1275 bolt objects are annotated and the number of each category is given. It can be found that in NPU-BOLT dataset, most of bolt objects are categorized into bolt nut and bolt head. 
\begin{figure}[ht]
  \centering
  \includegraphics{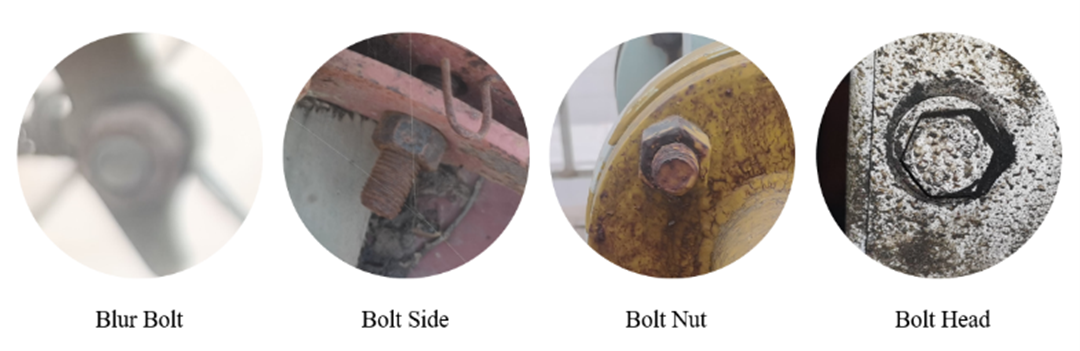}
  \caption{ Four categories in bolt target detection.}
  \label{fig:fig4}
\end{figure}
\begin{figure}[ht]
  \centering
  \includegraphics{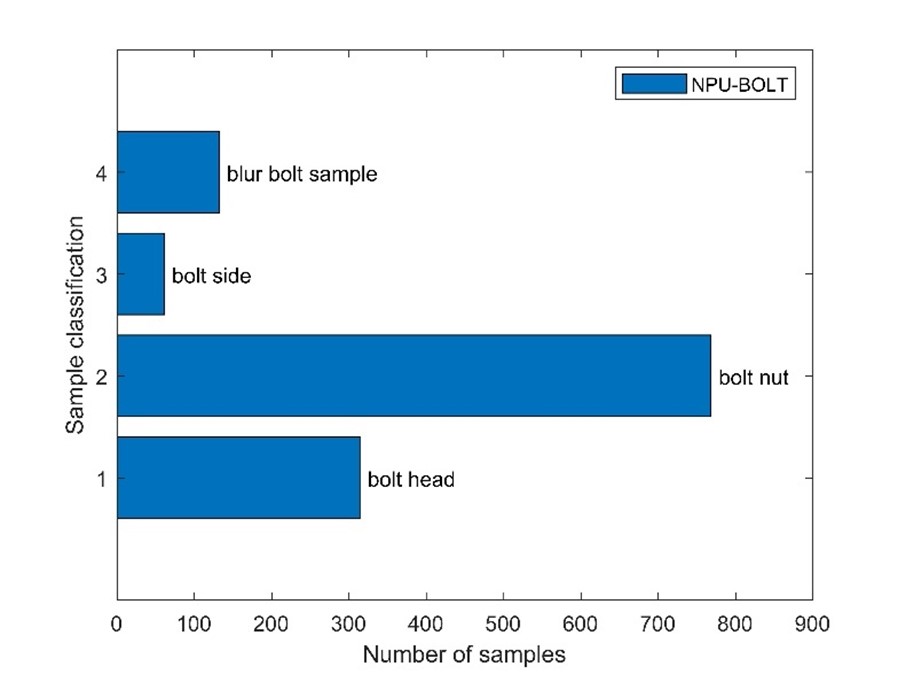}
  \caption{The number of each bolt objective category.}
  \label{fig:fig5}
\end{figure}
\subsection{Annotation method}
There are two common annotation methods. One is the bounding box annotation method and the other is the semantic segmentation (\citet{7}) pixel labeling. In this paper, to consider the annotation accuracy and application feasibly, the boundary box labeling method is chosen. 
The annotation tool uses the open source annotation software LabelImg (\citet{8}), which can output annotation file stored in XML format. XML also stores information of the size, name, path of the original image. It is important to note that the annotation data is recorded through the points (x1,y1,x2,y2) in the upper left and lower right corners of the bounding box.
The storage format of this dataset refers to the PASCAL VOC (\citet{9}) dataset format. The file is divided into three parts. Annotations folder stores the annotation information of the dataset. The ImageSets folder stores the index of the datasets used for training, validation, and testing. The JPEGImages folder stores the original images. The unified format allows the user to quickly deploy this dataset in his model.

\subsection{Dataset splits}
In order to ensure the accuracy of the model, 90\% image samples are randomly selected for training. Among them, 81\% was used as training set, 9\% as validation set, and the rest as test set.  During subsequent model training and testing, the test set will remain the same to ensure the accuracy of the evaluation of different models.

\section{Evaluations}

In this paper, the effectiveness of the NPU-BOLT dataset is evaluated by the latest target detection algorithms, which are  Faster-RCNN algorithm (\citet{10}), Centernet algorithm (\citet{11}) and YOLOv5 series algorithm (\citet{12}). The backbone of Faster-RCNN and Centernet is Resnet-50 (\citet{13}). The feature extraction capabilities of Resnet-50 in object detection models have been extensively verified.
In the yolov5 series of algorithms, there are four network structures: s, m, l, and x. Its network depth and complexity have gradually increased. Among these models, we chose the s model and the l model. The s model is the most portable network structure of the yolov5. Compared with the s model, the l model has higher detection accuracy.

\subsection{Training}
The training and prediction of this dataset will be carried out on the following platforms. The environment configuration is shown in Table 2.

\begin{table}[ht]
 \caption{Environment configuration}
  \centering
  \begin{tabular}{ll}
    \midrule
    \centering
        System &	Ubuntu 18.04\\
        GPU &	RTX 3090\\
        Pytorch version	 & 1.7.1\\
        Cuda version &	11.0\\

    \bottomrule
  \end{tabular}
  \label{tab:table2}
\end{table}
In addition, it should be noted that for each target detection algorithm, the selection of the pre-trained model initial parameters is very important. The pre-training models used in the training process in this paper are all from the VOC2007+2012 dataset. Meanwhile, in order to enhance the adaptability of the model, we refer to the mosaic data enhancement method proposed by the yolov5 series in the training set part, and apply this method to other object detection models.
In this paper, a training strategy of data augmentation, freeze parameters and multiple epochs are used.  The total number of epochs is set as 600.  Since the pre-training model has been fully trained on the VOC public dataset, in the initial 200 epochs, the backbone network is frozen. During the training process, the parameters of the backbone network are not changed, and only some of the parameters of the output layer are modified, such as Resnet50 in Faster-RCNN and CSPDarknet in yolov5.  During the next 400 epochs, the frozen parameters are unfrozen and trained.

\subsection{Predict}
NPU-BOLT dataset contains bolt image at different scales. Some images captured by UAV show the characteristics of high resolution but with very small target size. In order to ensure the stability of the algorithm's training and prediction process for such images, image segmentation on high-resolution images to ensure the accuracy of detection results is performed. Meanwhile, due to the short capture distance, some high-resolution images only contain a small number of bolt objects, and have a large relative size. In this case, if the image segmentation operation is adopted, the detection target will be lost. Therefore, the following strategies are used in the model prediction process.
For images larger than 2000*2000, they are divide into two channels. In the Channel 1, first selects a sliding frame of 900*900 to traverse the image and sets the step-size to 450 pixels to generate image data after image segmentation. Then the channel data is inputted into the network to obtain the target detection results of each segmented image. Thirdly, combine the results in the original image, delete the duplicate items, and generate the result graph of the image in channel 1. In the Channel 2, input the original image directly into the network to obtain the results and generate the resulting graph of the image. Finally, combine the results of channel 1 and 2, delete the duplicate items as the final output.

\subsection{Result}
After fully training each model, the object detection results are obtained.  The main index AP (Average Precision) value is shown in Table 3. It is noted that confidence thresholds are both 0.5.

\begin{table}[ht]
 \caption{The AP performance of each object detection model for the NPU-BOLT dataset}
  \centering
  \begin{tabular}{lllll}
    \toprule
           & YoloV5-s	&YoloV5-l&	\makecell[c]{Faster-RCNN\\(resnet50)}&\makecell[c]{CenterNet\\(resnet50)}\\
    \midrule
    \centering
    Bolt head & 93.77\%	& 97.38\% & 92.37\% &79.17\%\\
    Bolt nut  &	81.21\% & 91.88\% & 86.23\%	&84.54\%\\
    Bolt side &	0\%     & 0\%     &	1.51\%  &0\%\\
    Blur bolt &	0\%     & 0\%     &	4.13\%  &7.45\%\\
    \bottomrule
  \end{tabular}
  \label{tab:table3}
\end{table}
From the result in Table 2, it can be found that the classification ability of the model on the four categories is quite different. The two main object types of the dataset, bolt head and bolt nut, can be well identified and classified. However, for bolt side and blur bolt sample targets, the location and classification cannot be well obtained. 
It must be noted that using the developed NPU-BOLT dataset together with common popular object detection algorithms, bolt feature extraction and object detection can be achieved even under very complex natural environment conditions. It shows that the dataset can be used for training popular deep learning algorithms for the aim of bolt object detection in natural scenes. 
For bolt side and blur samples, the current object detection model cannot achieve good positioning and classification. The main problem is the following three points. First, the image of the dataset in this paper is mainly consisting of bolt nut and bolt head categories. The number of bolt side samples and blurred samples are not sufficient. There is no complete feature information for training. Second, the bolt side often contains other associative features such as threads, which increase the difficulty of identification. Meanwhile, due to the influence of the capture direction, the bolt side is not suitable for the traditional detection box labeling method. It is more suitable for the label box model with rotation angle. Third, blur samples are also a major problem in the field of current object detection. 
Finally, the detail performance index for yolov5-l model is given in Table 4. The reason it that among the four deep learning algorithm, the yolov5-l model performs the best.  
\begin{table}[h]
 \caption{The AP performance of each object detection model for the NPU-BOLT dataset}
  \centering
  \begin{tabular}{lllll}
    \toprule
        &AP	 &F1 &	recall &	precision\\
    \midrule
    \centering
    Bolt head &	97.38\%	& 0.94 &	88.64\%	 &99.15\%\\
    Bolt nut  &	91.88\% & 0.83 &	94.44\%	 &73.91\%\\
    Bolt side &	0\%	    & 0	   &0\%          &	0\%\\
    Blur bolt &	0\%	    & 0    &0\%          &	0\%\\

    \bottomrule
  \end{tabular}
  \label{tab:table4}
\end{table}

\section{Conclusion}
To largely promote the research on the bolt loosening detection method that using deep learning and machine learning technique and to make the method more practicality, a bolt dataset with images from real natural scenes is developed in this paper. The developed NPU-BOLT dataset consists of more than 300 bolt images capturing in natural scenes. As far as the authors know, the developed dataset is the first public dataset dedicated to bolt object detection using natural scenes images. 
The NPU-BOLT dataset provided in this paper has been uploaded publicly available on Kaggle web platforms. In the first version of the dataset, it contains 337 samples of bolt joints images mainly in the natural environment, with image data sizes ranging from 400*400 to 6000*4000, totaling approximately 1275 bolt targets. The bolt targets are annotated into four categories named blur bolt, bolt head, bolt nut and bolt side. The dataset is tested with advanced object detection models including yolov5, Faster-RCNN and CenterNet. The evaluation results show that the bolt target detection model trained using this dataset can well locate and classify the bolt head and bolt nut in natural environment. In the yolov5-l model, the average precision of the two main categories reach 97.38\% and 91.88\%, respectively. The proposed dataset bridges the gap in the current field of bolt object detection. Meanwhile, it is welcomed the related researchers supplement and improve the dataset in the future.

\bibliographystyle{plainnat}
\bibliography{references}  

\end{document}